\documentclass[letterpaper, 10 pt, journal, twoside]{IEEEtran}

\usepackage{graphicx}
\usepackage{amsmath}
\usepackage{amssymb}
\usepackage{array}
\usepackage{url}
\usepackage[capitalise]{cleveref}
\crefname{equation}{}{}

\ifCLASSINFOpdf
\else
\fi
\begin{document}
%
\title{Learning Maximal Safe Sets Using Hypernetworks for MPC-based Local Trajectory Planning in Unknown Environments}
%
%
%

\author{Bojan Derajić$^{1, 2}$, Mohamed-Khalil Bouzidi$^{1, 3}$, Sebastian Bernhard$^{1}$ and Wolfgang Hönig$^{2}$
\thanks{Manuscript received: March 1, 2025; Revised: May 27, 2025; Accepted: June 29, 2025.}
\thanks{This paper was recommended for publication by Editor Clement Gosselin upon evaluation of the Associate Editor and Reviewers' comments.
This work was funded by the German Federal Ministry for Economic Affairs and Climate Action within the project \textit{nxtAIM}.} 
\thanks{$^{1}$Continental Automotive Technologies GmbH, Germany}%
\thanks{$^{2}$Technical University of Berlin, Germany}%
\thanks{$^{3}$Free University of Berlin, Germany}%
\thanks{Contact e-mail: {\tt\small bojan.derajic@continental.com}}%
\thanks{Digital Object Identifier (DOI): see top of this page.}
}
%
%

\markboth{IEEE Robotics and Automation Letters. Preprint Version. Accepted June, 2025}
{Derajić \MakeLowercase{\textit{et al.}}: Learning Maximal Safe Sets Using Hypernetworks for MPC-based Local Trajectory Planning} 

%



\maketitle

\begin{abstract}
This paper presents a novel learning-based approach for online estimation of maximal safe sets for local trajectory planning in unknown static environments. The neural representation of a set is used as the terminal set constraint for a model predictive control (MPC) local planner, resulting in improved recursive feasibility and safety. To achieve real-time performance and desired generalization properties, we employ the idea of hypernetworks. We use the Hamilton-Jacobi (HJ) reachability analysis as the source of supervision during the training process, allowing us to consider general nonlinear dynamics and arbitrary constraints. The proposed method is extensively evaluated against relevant baselines in simulations for different environments and robot dynamics. The results show an increase in success rate of up to 52\% compared to the best baseline while maintaining comparable execution speed. Additionally, we deploy our proposed method, NTC-MPC, on a physical robot and demonstrate its ability to safely avoid obstacles in scenarios where the baselines fail.
\end{abstract}


%
\IEEEpeerreviewmaketitle

\section{Introduction}

\IEEEPARstart{S}{afety} persists as one of the key challenges in the field of autonomous robotics. Real-world applications require modern robots to operate in unstructured and often unknown environments. At the same time, they must maintain rigorous safety standards during operation, including collision avoidance and satisfaction of actuation limits. The requirement to achieve desired performance and safety simultaneously motivated the development of many advanced algorithms for local motion planning \cite{zhang_optimization-based_2021, rosmann_timed-elastic-bands_2015, williams_aggressive_2016}.

Local motion planning becomes especially difficult in cluttered environments, when the robot's actuation is significantly limited or when the task imposes additional restrictions on mobility. Actuation capabilities are limited by design, while applications such as mobile manipulation or transportation of heavy and unstable objects require very smooth motion of the mobile base. All such conditions together constrain the feasible states and motions for a robot in order to fulfill its goal.

In general, the goal is to reach a desired state while preventing the robot from entering the \textit{failure set}, which usually represents collision or violation of some other constraints. Consequently, this requires avoiding all states that would eventually lead to the failure set despite any control actions, which is a well-known problem in mobile robotics \cite{fraichard_short_2007, hsu_safety_2024, hwang_safe_2024}. In the context of Hamilton-Jacobi (HJ) reachability analysis, all states that have to be avoided form a backward reachable tube (BRT) \cite{bansal_hamilton-jacobi_2017}, whereas the complement of the BRT is the \textit{maximal safe set}.

\begin{figure}[t]
    \centering
    \includegraphics[width=0.8\columnwidth]{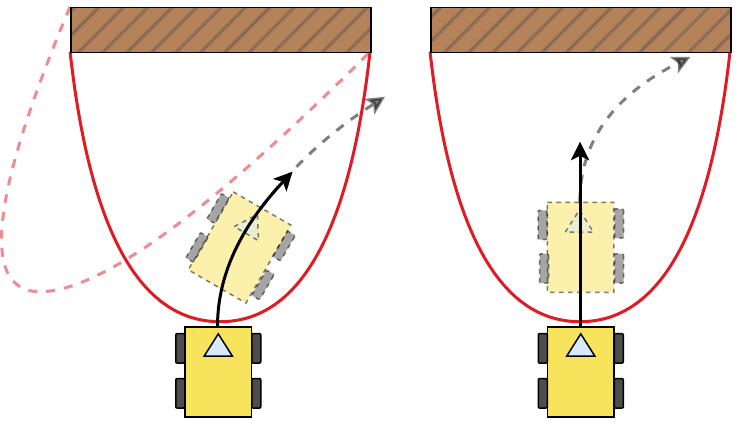}
    \caption{Obstacle avoidance with an MPC local planner. The red curve marks the BRT boundary (solid line - current time; dashed line - future time), indicating inevitable collision. Left: The planner with BRT-based constraint avoids collisions regardless of horizon length. Right: The planner with distance-based constraint fails if the horizon is too short.}
    \label{fig:intro_illustration}
\end{figure}

Our work focuses on real-time estimation of maximal safe sets for obstacle avoidance in unknown static environments based on local observations. We efficiently integrate the resulting safe set representation with a model predictive control (MPC) local planner as the terminal set constraint. To obtain the learned estimator, we numerically calculate the true maximal safe regions via the HJ reachability framework for a set of local observations based on robot dynamics. Then, we use the dataset to train a model architecture consisting of a hypernetwork and a main network in a supervised fashion. The hypernetwork learns how to parameterize the main network for different observations while the main network renders the corresponding estimated maximal safe set in real-time, which we use as the terminal constraint for the MPC planner. Using a BRT-based constraint allows for safe obstacle avoidance even if the prediction horizon is not long enough to detect a collision, which is not true for a distance-based constraint as illustrated in \cref{fig:intro_illustration}.

Compared to the existing approaches based on HJ reachability, our method can approximate the HJ value function in real-time for a continual stream of local observations in unknown environments, which we illustrate for two different nonholonomic robot dynamics. To the best of our knowledge, none of the existing methods can be applied to such a scenario without additional simplifying assumptions. Therefore, the approach introduced in this paper brings a significant improvement in the applicability of HJ reachability for real-time systems. The main contributions of this paper are the following:
\begin{itemize}
    \item Learning-based approach for real-time estimation of maximal safe sets for general nonlinear systems based on local observations in unknown static environments. 
    \item A novel MPC architecture integrating approximated maximal safe sets as the neural terminal constraint for improved recursive feasibility.
    \item A custom loss function for regression tasks with improved zero-level approximation.
\end{itemize}

\section{Related Work}
\label{sec:related_work}

One of the core problems in local motion planning is designing an environment representation based on sensor data and integrating it within the planner in real-time. The representation has to provide sufficient information to safely avoid obstacles while progressing along the reference path. In practice, an Euclidean distance-based signed distance function (SDF) is a common choice due to its intuitive interpretation and efficient computation \cite{oleynikova_signed_2016}.

The primary disadvantage of an SDF is the failure to account for the robot dynamics and actuation limits. The safe set associated with the SDF is usually an overapproximation of the maximal safe set, leading to higher collision rates \cite{fraichard_short_2007, hwang_safe_2024}. An alternative is to use control barrier functions (CBFs), which are based on Nagumo's Theorem of invariance \cite{ames_control_2019}. CBFs are used for different safety-critical control tasks and a discrete-time CBF was applied as a safety constraint for MPC-based trajectory planning in \cite{zeng_safety-critical_2021}. Even though CBFs can provide desirable safety guarantees, designing a proper CBF for a general nonlinear dynamical system remains very challenging and the existing approaches usually fail to characterize the safe regions well, resulting in conservative or unsafe motion.

Computing the maximal safe set is possible using HJ reachability analysis, which is a framework for model-based computation of reachable sets and system verification \cite{bansal_hamilton-jacobi_2017}. This framework suffers from the curse of dimensionality and is impractical for high-dimensional systems. Improved scalability is achieved by efficient initialization \cite{herbert_reachability-based_2019}, system decomposition \cite{chen_decomposition_2018}, and neural approximation of the HJ solution \cite{bansal_deepreach_2021}. However, those approaches assume complete \textit{a priori} knowledge about the environment, such as a global map, to compute the HJ value function. The obtained value function would then be stored in memory and queried online during motion planning, depending on the robot's position in the global map. If the global map is not available in advance or if the distribution of obstacles changes frequently, those methods have limited practical value.

To improve flexibility, the work in \cite{borquez_parameter-conditioned_2023} introduced parameter-conditioned reachable sets by augmenting system dynamics with "virtual" states representing environment-dependent parameters. This idea was applied for motion planning in mobile robotics \cite{jeong_parameterized_2024} and autonomous driving \cite{nakamura_online_2023}. The main limitation of this method is the ability to handle a relatively small number of environmental parameters, while our approach conditions the HJ solution based on high-dimensional observations. A different approach is used in \cite{ramesh_kumar_fast_2023} where the HJ value function is approximated numerically in a receding horizon fashion via the iterative linear-quadratic (iLQ) algorithm. Nevertheless, compared to our method, the iLQ approximation is valid only locally, the hard constraints in the system are not handled explicitly, and the resulting solution depends on initialization and system nonlinearity.  

In comparison with other methods for safe navigation in unknown environments, the main advantage of our approach is improved real-time performance. For example, the method in \cite{bajcsy_efficient_2019} uses the HJ value function as the safety constraint, but requires $\sim$600 ms to find a solution in the best case. Authors in \cite{lafmejani_nmpc-lbf_2022} generate a neural SDF and use it as a CBF constraint for an MPC planner, requiring 150 ms for weights update and 250 ms for optimization. In contrast, depending on the horizon length and robot model, our method takes between 5 and 13 ms on average. Also, compared to MPC planners with neural constraints for collision avoidance \cite{jacquet_n-mpc_2024} and system dynamics \cite{salzmann_real-time_2023}, the proposed approach uniquely employs a neural model as the MPC terminal constraint. 

\section{Preliminaries}
\label{sec:preliminaries}

\subsection{Problem Formulation}
\label{subsec:problem_formulation}

We consider a nonlinear dynamical system
\begin{equation} \label{eq:general_dynamics}
    \dot{x}(t) = f(x(t), u(t)),
\end{equation}
where $x(t) \in \mathbb{R}^n$ represents system's state vector, $u(t) \in \mathbb{R}^m$ is the control vector, and $t \in \mathbb{R}$ denotes the time variable. The discrete-time form of \cref{eq:general_dynamics} is
\begin{equation}
    x_{k+1} = f_d(x_k, u_k),
\end{equation}
where $k \in \mathbb{Z}$ is the discrete-time variable, $x_k$ and $u_k$ are state and control vectors at time $t=k \cdot \delta t$ for a fixed sampling time $\delta t$, and $f_d$ is discrete-time model obtained by some discretization method from the continuous-time dynamics $f$.

Since our aim is to employ an MPC local planner, we first define the corresponding optimal control problem that is solved iteratively at every time step $k$:
\begin{subequations}
\begin{align}
    \min_{u_{k:k+N-1}} \quad &p(x_{k + N}) + \sum_{i = 0}^{N - 1} q(x_{k+i}, u_{k+i}) \label{eq:opt_prob} \\
    s.t. \quad &x_{k+i+1} = f_d(x_{k+i}, u_{k+i}), \, i=0:N-1 \label{eq:dynamics_constr} \\
    &x_{k+i} \in \mathcal{X}, \quad i=0:N \label{eq:state_constr} \\
    &u_{k+i} \in \mathcal{U}, \quad i=0:N-1 \label{eq:control_constr} \\
    &h_{k+i}(x_{k+i}) \geq 0, \quad i=0:N \label{eq:safety_constr}
\end{align}
\end{subequations}
In this formulation, $x_{k+i}$ and $u_{k+i}$ are predictions of the state and control vectors at future time $k+i$ made at the current time step $k$ over the horizon of length $N$. Equality constraint \cref{eq:dynamics_constr} enforces the optimal solution to satisfy the system's dynamics, \cref{eq:state_constr} and \cref{eq:control_constr} constrain predicted state and control vectors to have admissible values, while \cref{eq:safety_constr} is an inequality constraint that incorporates any additional requirements imposed on the system states. The cost terms $p$ and $q$ depend on a specific task and in this paper we use the standard quadratic form: 
\begin{equation} \label{eq:q}
    q(x, u) = (x - x_r)^\top Q (x - x_r) + u^\top R u, 
\end{equation}
\begin{equation} \label{eq:p}
    p(x) = (x - x_r)^\top Q_N (x - x_r),
\end{equation}
where $x_r$ is the reference state and $R$, $Q$ and $Q_N$ are weight matrices. We focus our attention on ${h: \mathbb{Z} \times \mathbb{R}^n \rightarrow \mathbb{R}}$ in \cref{eq:safety_constr}. The main problem addressed in this paper is how to design function $h$ in real-time such that its zero-superlevel set approximates the maximal safe region.

\subsection{Hamilton-Jacobi Reachability Analysis}
\label{subsec:hj_reachability}

HJ reachability analysis is a model-based verification method used for dynamical systems under constraints and external disturbances. More formally, HJ reachability allows us to compute a BRT, which is a set of states such that if a system's trajectory starts from this set, it will eventually end up inside some failure set \cite{bansal_hamilton-jacobi_2017, bansal_deepreach_2021}. In the context of local motion planning for mobile robots, one can think of a BRT as a region around obstacles from which collision will happen inevitably at some point in time (e.g., due to long braking distance, limited steering, or short prediction horizon).

In this paper, we consider undisturbed dynamics described by \cref{eq:general_dynamics}. If we assume that the system starts at state $x(t)$, then we denote with $\xi^{u}_{x, t}(\tau)$ the system's state at time $\tau$ after applying $u(\cdot)$ over time horizon $[t, \tau]$. Here, we assume that $f$ satisfies standard assumptions of the existence and uniqueness of the state trajectories. We denote the failure set as $\mathcal{L}$ and define BRT for which the system will reach the set $\mathcal{L}$ within the time horizon $[t, T]$ as
\begin{equation}
    \mathcal{V}(t) = \{ x: \forall u(\cdot), \, \exists \tau \in [t, T], \, \xi^{u}_{x, t}(\tau)  \in \mathcal{L} \}.
\end{equation}

To compute BRT for a given failure set, we define $\mathcal{L}$ as the zero-sublevel set of a failure function $l(x)$, i.e. $\mathcal{L} = \{x:l(x) \leq 0 \}$. After that, the computation of the BRT is formulated as an optimization problem that seeks to find the minimal distance to the set $\mathcal{L}$ over the time horizon, i.e.
\begin{equation}
    J(x, t, u(\cdot)) = \min_{\tau \in [t, T]} l(\xi^{u}_{x, t}(\tau)).
\end{equation}
Since $\mathcal{L}$ represents an unsafe region, the goal is to find the optimal control that will maximize this distance. Therefore, we introduce the value function corresponding to this optimal control problem:
\begin{equation} \label{eq:value_func}
    V(x, t) = \sup_{u(\cdot)} \{ J(x, t, u(\cdot)) \}.
\end{equation}
The value function is computed numerically over a discrete state-space grid using the dynamic programming principle,  resulting in the final value Hamilton-Jacobi-Isaacs Variational Inequality (HJI VI) \cite{bansal_deepreach_2021, barron_bellman_1989}:
\begin{equation} \label{eq:hji_vi}
    \begin{aligned}
    \min \left\{ \frac{\partial}{\partial t}V(x, t) + H(x, t), \, l(x) - V(x, t) \right\} &= 0, \\
    V(x, T) &= l(x).
    \end{aligned}
\end{equation}
In the formulation above, Hamiltonian $H(x, t)$ is defined as
\begin{equation}
    H(x, t) = \max_{u} \nabla V(x, t) \cdot f(x, u).
\end{equation}
Once the value function \cref{eq:value_func} is calculated, the corresponding BRT can be obtained as its zero-sublevel set, i.e.
\begin{equation}
    \mathcal{V}(t) = \left\{ x: V(x, t) \leq 0 \right\},
\end{equation}
and the maximal safe set is its complement.

\section{Methodology}
\label{sec:methodology}

The main goal of our work is to design the constraint function \cref{eq:safety_constr} online based on available local observations of the environment. Since we primarily focus on local motion planning, we assume that \cref{eq:safety_constr} represents collision avoidance constraint. A direct choice would be a distance-based constraint such as SDF observed at the current time. However, if the prediction horizon is not long enough to perceive collision by the end of the horizon, the robot might enter the BRT of the obstacle which would lead to infeasibility and inevitable collision later in time. Therefore, the constraint \cref{eq:safety_constr} should prevent the robot from entering the BRT at any point in time.

From the MPC literature, it is well known that the recursive feasibility can be achieved if the terminal set is control invariant, which is true for the maximal safe set\footnote{This set is also known as the \textit{viability kernel}.} introduced above \cite{kerrigan_invariant_2000, liniger_viability_2015, blanchini_survey_1999}. This motivates us to use the approximated maximal safe set as the terminal set constraint, while still using the observed SDF as the stage constraint. Nevertheless, computing maximal safe regions for general nonlinear systems under control and state constraints involves solving \cref{eq:hji_vi}, which is intractable in real-time. Therefore, we aim for ML techniques capable of mapping directly from environment observation to the approximation of the maximal safe set in the form of a parameterized model.

\subsection{Model Architecture}
\label{subsec:model_architecture}

The designed model has to satisfy two objectives: generalize well across a wide spectrum of possible observations and allow for computationally efficient trajectory optimization in real-time. Adequate generalization requires expressive models, i.e., models with a large number of trainable parameters optimized on a vast amount of data. On the other side, a computationally efficient representation requires a simple model architecture with a relatively small number of parameters. This feature is particularly important for local planners relying on iterative numerical optimization methods such as MPC.

Our approach separates the ML model into two neural networks. The first network processes the local representation of the environment (e.g., local SDF or costmap) and outputs the parameters of the second network, which represents the constraint function in \cref{eq:safety_constr}. In the literature, the first network is referred to as \textit{hypernetwork}, and the second network is known as \textit{main network} \cite{ha_hypernetworks_2017}. The hypernetwork is designed to be expressive and have a large number of trainable parameters to achieve desired generalization properties. Complementary, the main network is a simple and efficient model that approximates the maximal safe set corresponding only to the current observation of the environment. During motion planning, the hypernetwork is inferred only once per time step to parametrize the main network which is then provided to the optimization algorithm as a constraint function and it is iteratively queried for its value and gradient. Moreover, the main network is naturally differentiable via backpropagation and does not require gradient approximation. As shown in \cite{galanti_modularity_2020}, the total number of trainable parameters can be orders of magnitude smaller compared to the embedding-based approach. Additionally, in \ref{subsec:single_net_architecture} we analyzed the potential use of a single-network architecture to justify the benefits of the hypernetworks for real-time performance.

\begin{figure}[t]
    \centering
    \includegraphics[width=\columnwidth]{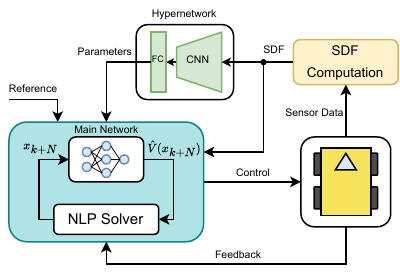}
    \caption{Architecture of the proposed NTC-MPC local planner. An SDF computed from local observation is used as the MPC stage constraint and hypernetwork input. The hypernetwork parametrizes the main network to approximate the maximal safe set locally and is used as the MPC terminal set constraint.}
    \label{fig:full_diag}
\end{figure}

In our concrete implementation, we assume that the local observation is provided as an SDF. This is a common representation used in practice for mobile robots and can be efficiently computed, e.g., from an occupancy grid map. Since the input is in the form of a single-channel image, our hypernetwork consists of a deep convolutional neural network (CNN) backbone and a fully connected (FC) head. After the parameters are obtained, the main network is used as the terminal constraint, which approximates the value function in \cref{eq:value_func} and implicitly the maximal safe set. The only assumption is that the costmap is large enough so that the obstacle is visible before the robot enters the corresponding BRT. As a main network, we use a simple multi-layer perceptron (MLP) model. The overall architecture of our approach is presented in \cref{fig:full_diag}. 

\subsection{Training Procedure}
\label{subsec:training_procedure}

The hypernetwork is trained in a supervised fashion using input data and ground truth labels. The input in our implementation is a set of 2D SDFs and a set of state-space grid points for the concrete robot dynamics. We can obtain SDFs from a simulation or real-world environments, while the state-space grid points are generated by a uniform discretization of the state space depending on the desired resolution and state limits.

\begin{figure}[t]
    \centering
    \includegraphics[width=\columnwidth]{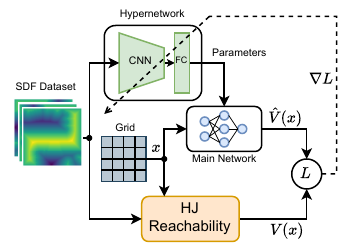}
    \caption{Training diagram for the proposed model architecture. A set of SDFs and state-space grid points represent input data, while the HJ reachability analysis is used to obtain output labels. Only the hypernetwork contains trainable parameters, which are optimized through the gradient descent procedure.}
    \label{fig:train_diag}
\end{figure}

We utilize HJ reachability to calculate the true value function in \cref{eq:value_func} over the state-space grid for a theoretically infinite-time horizon\footnote{In practice, we propagate the value function until it converges in time. Also, since the value function in this case corresponds to the infinite-time horizon, we exclude the time variable from it in the rest of the paper.} and use those values as the labels for the predicted value function $\hat{V}(x)$ at the output of the main network. To compute the true value function $V(x)$, the failure function $l(x)$ is needed as the boundary constraint in \cref{eq:hji_vi}. Since our failure region represents occupied space in the locally observed environment, we define $l(x)$ as the SDF corresponding to the local observation. The overall training procedure is illustrated in \cref{fig:train_diag}.

To improve maximal safe set approximation additionally, we propose a new loss function instead of the standard mean squared error (MSE) loss. The MSE loss penalizes prediction errors equally over the whole state-space grid. However, the main purpose of the learned value function is to separate safe and unsafe regions by its zero-level set. Also, the importance of the set boundary is justified by Nagumo’s Theorem of invariance \cite{ames_control_2019}. Therefore, we introduce a radially-weighted MSE (RWMSE) loss that favors higher prediction accuracy near the zero-level set of the true value function:
\begin{equation} \label{eq:rwmse}
    \begin{aligned}
        L &= \frac{1}{M} \sum_{i=1}^{M} \frac{1}{K} \sum_{j=1}^{K} w_{i, j} (V_{i, j} - \hat{V}_{i, j})^2, \\
        w_{i, j} &= 1 + \alpha \exp \left( -\beta V_{i, j}^2 \right), \quad \alpha, \beta \in \mathbb{R}^+.
    \end{aligned}
\end{equation}
In the definition above, $M$ is the total number of training samples (SDFs), $K$ is the number of states in the state-space grid, while $\alpha$ and $\beta$ represent hyperparameters that control the weighting of individual grid points. The value of $\alpha$ determines the relative weighting of the prediction error near the zero-level set, while the value of $\beta$ controls the width of the margin around the zero-level set that is weighted higher. The proposed RWMSE empirically improves the characterization of the maximal safe sets and results in enhanced safety. 

\section{Experimental Results}
\label{sec:experimental_results}

\subsection{Robot Dynamics}
\label{subsec:robot_dynamics}

\textit{Dubins Car:} 1st-order unicycle model with constant linear speed and limited angular speed. In practice, this model describes systems for which linear speed is not controllable or its change over time is negligible. Some examples include fixed-wing aircraft, missiles, marine vessels, vehicles with faulty braking systems, etc. The state vector ${x = [x_p, \, y_p, \, \theta]^\top}$ includes positional coordinates $(x_p, y_p)$ and orientation angle $\theta$, while the control input is only the angular speed, i.e. ${u = \omega}$. The dynamics equation is
\begin{equation} \label{eq:dubins_car}
    \dot{x} = \left[
        v \cos(\theta), \,
        v \sin(\theta), \,
        \omega
    \right]^\top,
\end{equation}
where linear speed $v = 0.5$ m/s and control limit is ${\omega \in [-0.25, 0.25]}$ rad/s.

\textit{Dynamic Unicycle:} 2nd-order unicycle model used in practice to model differential drive and skid-steer types of robots. The state vector is ${x = [x_p, \, y_p, \, \theta, \, v, \, \omega]^\top}$, where $(x_p, y_p)$ are positional coordinates, $\theta$ is orientation, and $v$ and $\omega$ are linear and angular velocities, respectively. The control vector ${u = [a, \, \alpha]^\top}$ consists of linear acceleration $a$ and angular acceleration $\alpha$. The system dynamics is 
\begin{equation}
    \dot{x} = \left[
        v \cos(\theta), \,
        v \sin(\theta), \,
        \omega, \,
        a, \,
        \alpha
    \right] ^ \top.
\end{equation}
For this model, we assume the following state and control constraints: ${v \in [-1.0, 1.0]}$ m/s, ${\omega \in [-0.5, 0.5]}$ rad/s, ${a \in [-0.25, 0.25]}$ m/s$^2$, and ${\alpha \in [-1.0, 1.0]}$ rad/s$^2$.

\subsection{Model Training}
\label{subsec:model_training}

The model architecture described in \cref{sec:methodology} is implemented using PyTorch. The hypernetwork is implemented as a CNN model with a single-channel input and the main network is implemented as an MLP model. The number of outputs of the hypernetwork was equal to the number of parameters of the main network. For the Dubins car model, the main network has 3 inputs, 9 hidden layers with $[32, 32, 32, 16, 16, 16, 8, 8, 8]$ neurons, and a single output, resulting in 3601 parameters in total. The first 3 layers use sinusoidal activation, while the remaining hidden layers use the SELU activation function.

\begin{figure}[t]
    \centering
    \includegraphics[width=\columnwidth]{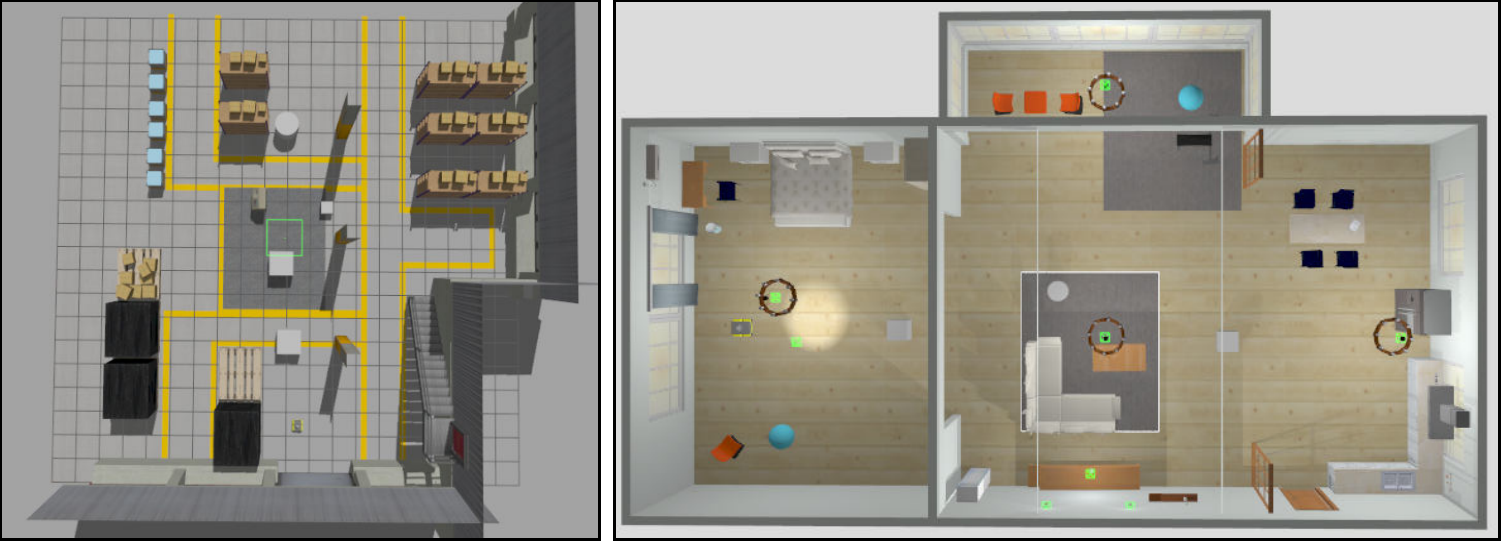}
    \caption{Two different Gazebo simulation environments. Left: Warehouse environment model. Right: House environment model.}
    \label{fig:sim_environments}
\end{figure}

The training dataset is created by first collecting a set of randomly placed local occupancy grid maps from the warehouse environment (see \cref{fig:sim_environments}) and then computing corresponding SDFs, which are used as input to the hypernetwork. We initially collected 2.500 occupancy maps of size 6$\times$6 m (100$\times$100 cells with resolution 0.06 m). To augment the dataset, we perform rotations for 90, 180 and 270 degrees to every original occupancy map and to its vertically flipped version, resulting in 20.000 samples in total. Then we define the state-space grid of size 100$\times$100$\times$20 and compute the true value function over that grid with the hj\_reachability\footnote{\url{https://github.com/StanfordASL/hj_reachability}} library.

\begin{figure}[b]
    \centering
    \includegraphics[width=\columnwidth]{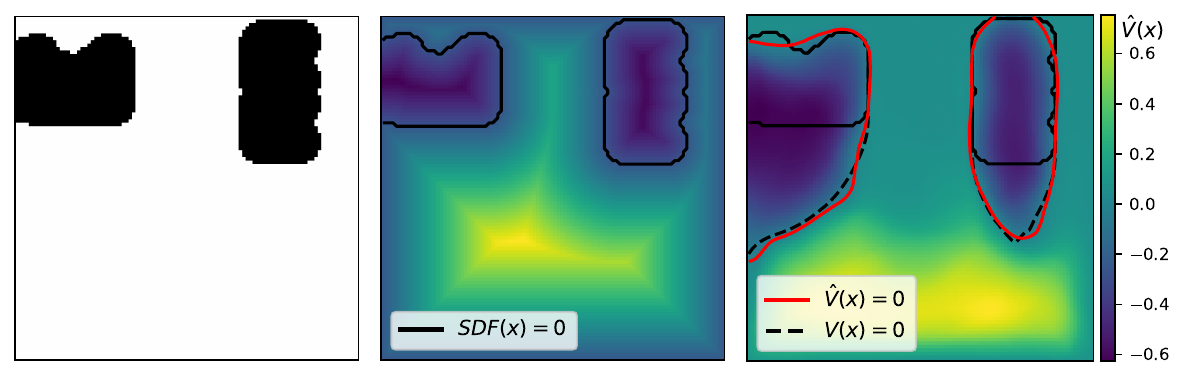}
    \caption{Results of the model training for the Dubins car model. Left: Local occupancy grid map. Middle: The corresponding SDF. Right: Learned value function slice for $\theta=0$ rad (robot is moving up) with predicted BRT boundary (solid red line) and the true BRT boundary (dashed black line).}
    \label{fig:ogm_sdf_vf}
\end{figure}

During the training process, an SDF is propagated through the hypernetwork to generate parameters that are loaded into the main network, and the complete state-space grid is propagated through the main network to estimate the HJ value function over the grid. The loss function is computed based on the true and estimated value functions and its gradient is used to update the parameters of the hypernetwork (see \cref{fig:train_diag}). In our experiments, we use 80\% of data for training and 20\% for validation. We train the hypernetwork model of $\sim 4.8 \times 10^6$ parameters for 150 epochs using Adam optimizer with the RWMSE loss defined in \cref{eq:rwmse}. Based on a grid search, the RWMSE hyperparameters are set to $\alpha = 1000$ and $\beta = 10$. For the batch size of 32, the training process takes about 7.5 hours on a single Nvidia RTX3090 GPU. \cref{fig:ogm_sdf_vf} shows the results of the model training for a sample environment observation. 

\begin{figure}[thpb]
    \centering
    \includegraphics[width=\columnwidth]{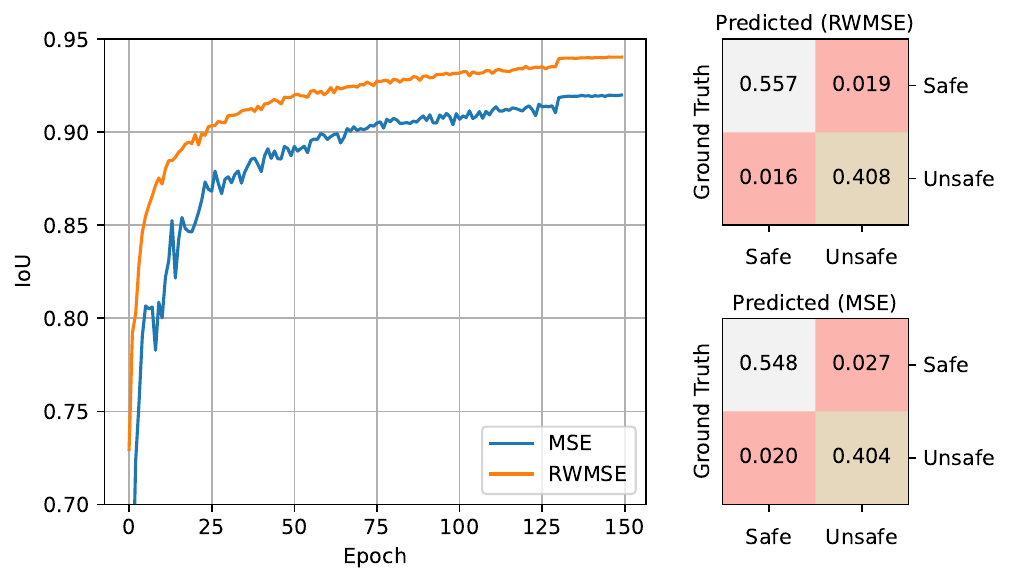}
    \caption{Evaluation of models trained for the RWMSE or MSE loss function for the Dubins car model. Left: IoU metric evaluated during training on the validation dataset. Right up: Confusion matrix for RWMSE loss. Right down: Confusion matrix for MSE loss.}
    \label{fig:mse_vs_rwmse}
\end{figure}

To examine the RWMSE loss function, we repeat the training process with MSE loss under exactly the same training settings and weights initialization. Since the values of the two loss functions are not directly comparable, we compare the Intersection over Union (IoU) metric for the safe region of the state space, which tells us how well the model recovers the maximal safe set. Additionally, to get more detailed insight into the results, we compare confusion matrices for models trained for the two losses. The results for the validation dataset are presented in \cref{fig:mse_vs_rwmse} and clearly show the advantages of the proposed RWMSE. 

For the dynamic unicycle model, we perform the training process analogously using the same model architectures, except that the main network has 5 inputs. The true HJ value function is computed over a state-space grid of size 50$\times$50$\times$7$\times$7$\times$7. However, due to increased data size, the model is trained with a batch size of 16 for 100 epochs with the same optimizer and loss function for 18 h. 

The hypernetwork inference time on the GPU is about $\sim$2 ms in both cases. In contrast, computing the HJ value function takes $\sim$1 s for the Dubins car and $\sim$2 s for the dynamic unicycle on the same hardware, which is an improvement of three orders of magnitude in execution speed. 

\subsection{Monte Carlo Simulations}
\label{subsec:mc_simulations}

After model training, we perform 3D simulations using Gazebo for both robot dynamics in two different environments shown in \cref{fig:sim_environments}. The models are trained only on the data from the warehouse environment, while the house environment is used to test the generalization properties and robustness to the distribution shifts. The two main metrics that we analyzed in our experiments are the success rate and optimization time of the MPC planner.

As the relevant baselines, we examine MPC planners with two other representations of the constraint in \cref{eq:safety_constr}. The first baseline uses the SDF observed at the time step $k$ 
\begin{equation} \label{eq:sdf_mpc}
    h_{k+i}(x_{k+i}) = h_{SDF, k}(x_{k+i}),
\end{equation}
 for $i=0, 1, \ldots, N$ and we refer to it as SDF-MPC. The second baseline called DCBF-MPC uses a safety constraint as given in \cite[Example A]{zeng_safety-critical_2021}, which represents a discrete-time CBF. This method defines the safety constraint as
\begin{equation} \label{eq:dcbf_mpc}
    h_{k+i}(x_{k+i}) = \Delta h_{SDF, k}(x_{k+i}) + \gamma h_{SDF, k}(x_{k+i}),
\end{equation}
for $i=1, 2, \ldots, N$ where $\Delta h_{SDF, k}(x_{k+i}) = h_{SDF, k}(x_{k+i}) - h_{SDF, k}(x_{k+i-1})$ and $\gamma \in (0, 1]$. Our method, as explained in \cref{sec:methodology}, uses the current SDF as the stage constraint, i.e. it uses \cref{eq:sdf_mpc} for $i=0,1, \ldots, N-1$, while the terminal constraint is defined by the main network
\begin{equation} \label{eq:ntc_mpc}
    h_{k+N}(x_{k+N}) = h_{main\_net}(x_{k+N}).
\end{equation}
We refer to our method as the \textit{Neural Terminal Constraint} MPC (NTC-MPC) local planner.

All planners are implemented using the CasADi framework \cite{andersson_casadi_2019} with IPOPT as NLP  solver and deployed as a ROS 2 package. The robot dynamics are discretized using the forward Euler method with a sampling time $\delta=0.1$ s. Additionally, we use the CasADi feature to autogenerate and compile C code before deployment. In the case of NTC-MPC, the hypernetwork is implemented as a PyTorch model, while the main network is implemented using CasADi functions so it can be efficiently integrated into the computational graph during optimization. Also, the parameters of the main network are treated as time-varying parameters of the underlying NLP problem, which is solved at every MPC iteration. Once a new SDF is available, the hypernetwork is inferred to estimate those parameters, and they are directly passed to the NLP solver along with the SDF, estimated state, and the reference state.

\begin{figure}[t]
    \centering
    \includegraphics[width=\columnwidth]{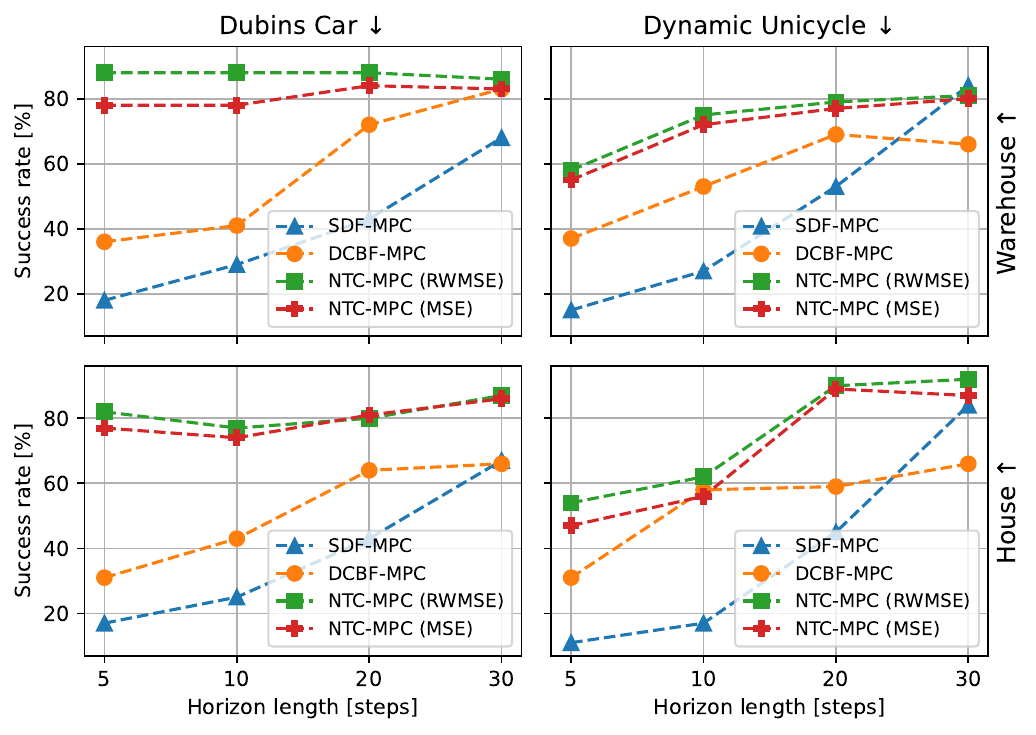}
    \caption{Success rates obtained in the simulation experiments for different environment models and robot dynamics. The performance of the baselines degrades when the prediction horizon is reduced, while our approach consistently maintains better performance.}
    \label{fig:success_rates}
\end{figure}

For both environments and robot models, we perform 100 simulations with randomly placed obstacles of different sizes and shapes over the environments. The experiments are repeated for different MPC methods and different horizon lengths. Moreover, to examine the impact of the proposed RWMSE loss on the success rate, we test the NTC-MPC method both for MSE and RWMSE loss functions. The task of the robot is to navigate from the initial position to the goal position based only on the local observations. This scenario is common when the two successive waypoints are provided by a global planner and the robot has to navigate between them using the local planner. An experiment is considered a failure if the robot collides with an obstacle or gets stuck. The obtained success rates are presented in \cref{fig:success_rates} and clearly illustrate better performance of the proposed method. As expected from the theoretical discussion, the NTC-MPC works well even for short prediction horizons, and the results show that the RWMSE loss function indeed has a positive impact on the success rates. Besides that, one could notice that the SDF-MPC method improves significantly when increasing the prediction horizon. However, this trend is ultimately bounded by the size of the local costmap, and increasing the MPC horizon further would not bring considerable improvement without increasing the robot's perception field.

\begin{figure}[htb]
    \centering
    \includegraphics[width=\columnwidth]{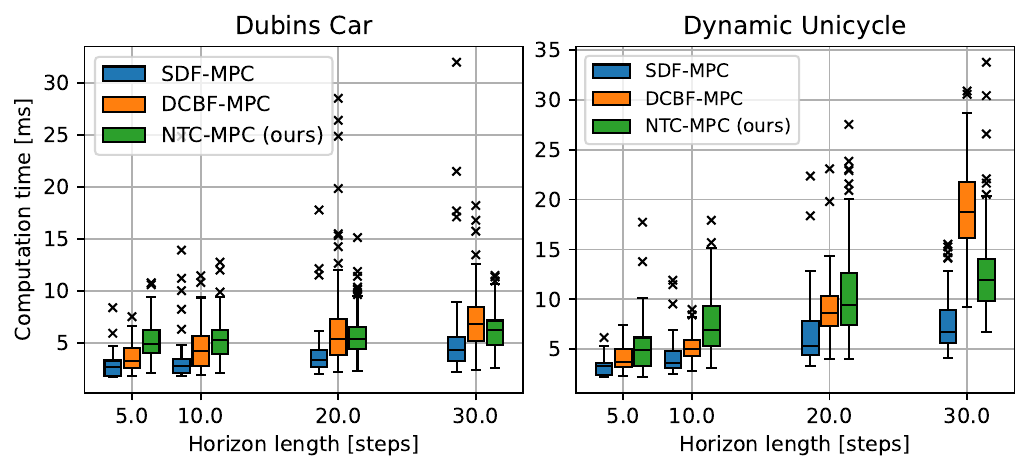}
    \caption{Computation times for different MPC local planners. Our method achieves real-time performance comparable to the baselines.}
    \label{fig:comp_time}
\end{figure}

Computation times reported in \cref{fig:comp_time} show that our NTC-MPC method achieves real-time performance comparable to the baselines. Moreover, for horizon lengths greater than 20 steps, it is even faster than the DCBF-MPC baseline. Since we replace only the terminal constraint of the SDF-MPC with the neural network, it causes a constant offset in the computation time compared to the SDF-MPC, regardless of horizon length. On the other hand, DCBF-MPC uses constraints based on SDF increment, which is more complex than the SDF function alone, and therefore the optimization complexity grows faster when increasing the horizon length.  

\subsection{Real-World Experiments}
\label{subsec:hardware_experiments}

In addition to the simulation experiments, we perform a set of collision avoidance experiments with a physical robot. For the experiments, we use a \textit{Continental} delivery mobile robot which is modeled as the Dubins car described by \cref{eq:dubins_car} in \cref{subsec:robot_dynamics}. The robot uses a 3D LiDAR for localization and local costmap computation. All computation is performed onboard with a mini PC Intel NUC11PHi7 (CPU: i7-1165G7 8x2.80GHz; RAM: 32GB; GPU: RTX2060 6GB).

\begin{figure}[tb]
    \centering
    \includegraphics[width=\columnwidth]{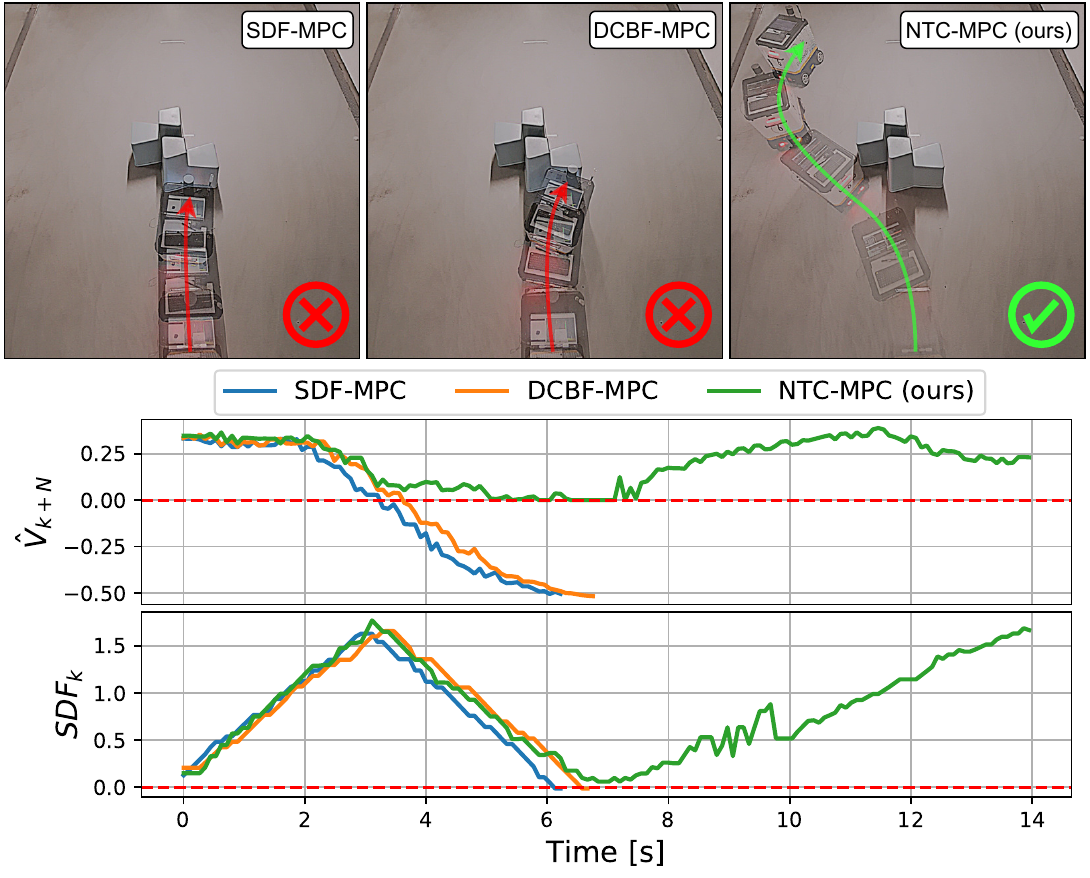}
    \caption{Robot motion and plots of $SDF(x_{k})$ and $\hat{V}(x_{k+N})$ for the Dubins car model and horizon length of 10 steps (1 s). Both baselines fail due to the short horizon, while our method successfully avoids collision. The plots illustrate that keeping $\hat{V}(x_{k+N}) > 0$ indeed prevents collision.}
    \label{fig:hardware_exp}
\end{figure}

We consider a task in which the local planner needs to avoid static obstacles between two waypoints based only on the local observations.  We perform experiments with different MPC methods and different horizon lengths for ten different distributions of static obstacles. The planners and the hypernetwork trained on synthetic data are directly transferred from the simulations to the real robot without any additional fine-tuning. The results of the experiments are presented in \cref{tab:hardware_exp}, while \cref{fig:hardware_exp} shows robot motion and plots of $SDF(x_{k})$ and $\hat{V}(x_{k+N})$ for the prediction horizon of 10 steps (1 s). The plots illustrate that keeping $\hat{V}(x_{k+N}) > 0$, i.e., keeping the terminal state $x_{k+N}$ within the maximal safe set, prevents collision. Our method achieves a higher success rate in all settings with comparable real-time performance, which is consistent with simulations and theoretical discussion.

\begin{table}[thpb]
    \caption{Results from Hardware Experiments with Dubins Car Model}
    \label{tab:hardware_exp}  
    \centering
    \renewcommand{\arraystretch}{1.5}
    \setlength{\tabcolsep}{10pt}
    \resizebox{\columnwidth}{!}{
    \begin{tabular}{|c|c|c|c|c||c|c|c|c|}
        \cline{2-9}
        \multicolumn{1}{c|}{} & \multicolumn{4}{c||}{Success Rate [\%]} & \multicolumn{4}{c|}{Mean Computation Time [ms]} \\
        \hline
        N [steps] $\rightarrow$ & 5 & 10 & 20 & 30 & 5 & 10 & 20 & 30 \\
        \hline
        \hline
        SDF-MPC & 0 & 0 & 30 & 70 & 2.65 & 4.55 & 6.26 & 6.54 \\
        DCBF-MPC & 0 & 10 & 70 & 80 & 4.06 & 5.61 & 5.91 & 7.76 \\
        NTC-MPC (ours) & \textbf{80} & \textbf{90} & \textbf{90} & \textbf{90} & 4.93 & 5.26 & 6.54 & 6.31 \\
        \hline
    \end{tabular}
    }
\end{table}

\subsection{Single-Network Architecture Analysis}
\label{subsec:single_net_architecture}

To emphasize the advantages of the proposed two-network architecture, we analyze the potential use of a single-network architecture and identify the practical difficulties that arise. Inspired by the work in \cite{borquez_parameter-conditioned_2023}, we consider an MLP model consisting of 3 hidden layers with 512 neurons. In this case, an SDF represents environment parameters that are treated as virtual states appended to the true states of the system. Since we use an SDF of size 100$\times$100, the joint state dimension exceeds $10 \times 10^3$, which is three orders of magnitude higher than the examples in \cite{borquez_parameter-conditioned_2023}. The total number of trainable parameters is $\sim5.6 \times 10^6$, which is the same order of magnitude as for the hypernetwork in our approach. 

From the computational efficiency aspect, integrating a neural constraint with such a large number of parameters in the optimization loop significantly increases optimization time, rendering the MPC planner impractical. For example, using such a model as the terminal constraint for the Dubins car and horizon length of 10 steps takes $\sim$1.6 s to find a solution, while our approach needs $\sim$5.5 ms on average.

Additionally, our proposed architecture is memory efficient during the training phase. For every SDF, we need to propagate the complete state-space grid through a model in order to obtain the complete value function at the output. Since the grid has a large number of points, the number of intermediate variables of the computational graph stored in RAM is also large. Concretely, to perform one forward and backward pass for one SDF in the case of the Dubins car model, the single-network model requires $\sim$9 GB of RAM, while our model occupies only  $\sim$0.7 GB of RAM.

\section{Discussion on Limitations}
\label{sec:discussion}

The main theoretical limitation of our method is the lack of safety guarantees, which is a common drawback of learning-based approaches \cite{lin_generating_2023}. Our major attempt in this direction is the proposed RWMSE loss function used during training, which improves the estimation accuracy in the vicinity of a BRT boundary. To achieve more desirable safety properties of the learned model, one could employ probabilistic measures of the approximation error to form chance constraints. Also, the HJ value function poses certain mathematical properties that could be exploited and embedded into the model architecture by custom design of the main network.

From the technical perspective, the major challenge arises from the curse of dimensionality associated with HJ reachability analysis during the data generation phase. The existing numerical tools are usually not applicable to systems with more than six state dimensions. To improve scalability, one can use existing techniques such as state decomposition \cite{chen_decomposition_2018} or neural solvers \cite{bansal_deepreach_2021} to generate data for higher-dimensional systems. Another direction would be to replace the supervised learning procedure with other types of learning. For example, using reinforcement learning (RL) as presented in \cite{fisac_bridging_2019} or self-supervision similar to \cite{bansal_deepreach_2021} are appealing options.

\section{Conclusions}
\label{sec:conclusion}

We present a learning-based method to approximate the HJ value function in real-time based on a high-dimensional observation of the environment, which is not possible with the existing methods in the field. Furthermore, we propose a new MPC architecture with a neural terminal constraint that uses the approximated maximal safe set as the terminal set constraint, resulting in improved recursive feasibility and safety, especially for short prediction horizons. We also introduce a custom RWMSE loss function, which achieves better results than the standard MSE loss.

For future work, we plan to extend the method to dynamic unknown environments by including additional layers of complexity, such as a predictive model of the environment and time-varying failure sets in HJ reachability. Also, our aim is to provide formal guarantees of safety through improved model architectures. Besides that, it would be interesting to apply the NTC-MPC method to applications other than mobile robotics that require the estimation of complex terminal sets in real-time.


\bibliographystyle{IEEEtran}
\bibliography{references}

\begin{thebibliography}{10}
\providecommand{\url}[1]{#1}
\csname url@samestyle\endcsname
\providecommand{\newblock}{\relax}
\providecommand{\bibinfo}[2]{#2}
\providecommand{\BIBentrySTDinterwordspacing}{\spaceskip=0pt\relax}
\providecommand{\BIBentryALTinterwordstretchfactor}{4}
\providecommand{\BIBentryALTinterwordspacing}{\spaceskip=\fontdimen2\font plus
\BIBentryALTinterwordstretchfactor\fontdimen3\font minus \fontdimen4\font\relax}
\providecommand{\BIBforeignlanguage}[2]{{%
\expandafter\ifx\csname l@#1\endcsname\relax
\typeout{** WARNING: IEEEtran.bst: No hyphenation pattern has been}%
\typeout{** loaded for the language `#1'. Using the pattern for}%
\typeout{** the default language instead.}%
\else
\language=\csname l@#1\endcsname
\fi
#2}}
\providecommand{\BIBdecl}{\relax}
\BIBdecl

\bibitem{zhang_optimization-based_2021}
X.~Zhang, A.~Liniger, and F.~Borrelli, ``Optimization-{Based} {Collision} {Avoidance},'' \emph{IEEE Trans. Control Syst. Technol.}, vol.~29, no.~3, pp. 972--983, 2021.

\bibitem{rosmann_timed-elastic-bands_2015}
C.~Rosmann, F.~Hoffmann, and T.~Bertram, ``Timed-{Elastic}-{Bands} for time-optimal point-to-point nonlinear model predictive control,'' in \emph{{IEEE} {Eur}. {Control} {Conf}. ({ECC})}, 2015, pp. 3352--3357.

\bibitem{williams_aggressive_2016}
G.~Williams, P.~Drews, B.~Goldfain, J.~M. Rehg, and E.~A. Theodorou, ``\BIBforeignlanguage{en}{Aggressive driving with model predictive path integral control},'' in \emph{\BIBforeignlanguage{en}{{IEEE} {Int}. {Conf}. {Robot}. {Autom}. ({ICRA})}}, 2016, pp. 1433--1440.

\bibitem{fraichard_short_2007}
T.~Fraichard, ``\BIBforeignlanguage{en}{A {Short} {Paper} about {Motion} {Safety}},'' in \emph{\BIBforeignlanguage{en}{{IEEE} {Int}. {Conf}. {Robot}. {Autom}. ({ICRA})}}, 2007, pp. 1140--1145.

\bibitem{hsu_safety_2024}
K.-C. Hsu, H.~Hu, and J.~F. Fisac, ``\BIBforeignlanguage{en}{The {Safety} {Filter}: {A} {Unified} {View} of {Safety}-{Critical} {Control} in {Autonomous} {Systems}},'' \emph{\BIBforeignlanguage{en}{Annu. Rev. Control Robot. Auton. Syst.}}, vol.~7, no. 2024, pp. 47--72, 2024.

\bibitem{hwang_safe_2024}
S.~Hwang, I.~Jang, D.~Kim, and H.~J. Kim, ``\BIBforeignlanguage{en}{Safe {Motion} {Planning} and {Control} for {Mobile} {Robots}: {A} {Survey}},'' \emph{\BIBforeignlanguage{en}{Int. J. Control Autom. Syst.}}, vol.~22, no.~10, pp. 2955--2969, 2024.

\bibitem{bansal_hamilton-jacobi_2017}
S.~Bansal, M.~Chen, S.~Herbert, and C.~J. Tomlin, ``Hamilton-{Jacobi} reachability: {A} brief overview and recent advances,'' in \emph{{IEEE} {Conf}. {Decis}. {Control} ({CDC})}, 2017, pp. 2242--2253.

\bibitem{oleynikova_signed_2016}
H.~Oleynikova, A.~Millane, Z.~Taylor, E.~Galceran, J.~Nieto, and R.~Siegwart, ``\BIBforeignlanguage{en}{Signed {Distance} {Fields}: {A} {Natural} {Representation} for {Both} {Mapping} and {Planning}},'' in \emph{\BIBforeignlanguage{en}{Workshop on {Geometry} and {Beyond}, {RSS}}}, 2016.

\bibitem{ames_control_2019}
A.~D. Ames, S.~Coogan, M.~Egerstedt, G.~Notomista, K.~Sreenath, and P.~Tabuada, ``Control {Barrier} {Functions}: {Theory} and {Applications},'' in \emph{{IEEE} {Eur}. {Control} {Conf}. ({ECC})}, 2019, pp. 3420--3431.

\bibitem{zeng_safety-critical_2021}
J.~Zeng, B.~Zhang, and K.~Sreenath, ``Safety-{Critical} {Model} {Predictive} {Control} with {Discrete}-{Time} {Control} {Barrier} {Function},'' in \emph{{IEEE} {Am}. {Control} {Conf}. ({ACC})}, 2021, pp. 3882--3889.

\bibitem{herbert_reachability-based_2019}
S.~L. Herbert, S.~Bansal, S.~Ghosh, and C.~J. Tomlin, ``Reachability-{Based} {Safety} {Guarantees} using {Efficient} {Initializations},'' in \emph{{IEEE} {Conf}. {Decis}. {Control} ({CDC})}, 2019, pp. 4810--4816.

\bibitem{chen_decomposition_2018}
M.~Chen, S.~L. Herbert, M.~S. Vashishtha, S.~Bansal, and C.~J. Tomlin, ``Decomposition of {Reachable} {Sets} and {Tubes} for a {Class} of {Nonlinear} {Systems},'' \emph{IEEE Trans. Autom. Control}, vol.~63, no.~11, pp. 3675--3688, 2018.

\bibitem{bansal_deepreach_2021}
S.~Bansal and C.~J. Tomlin, ``{DeepReach}: {A} {Deep} {Learning} {Approach} to {High}-{Dimensional} {Reachability},'' in \emph{{IEEE} {Int}. {Conf}. {Robot}. {Autom}. ({ICRA})}, 2021, pp. 1817--1824.

\bibitem{borquez_parameter-conditioned_2023}
J.~Borquez, K.~Nakamura, and S.~Bansal, ``Parameter-{Conditioned} {Reachable} {Sets} for {Updating} {Safety} {Assurances} {Online},'' in \emph{{IEEE} {Int}. {Conf}. {Robot}. {Autom}. ({ICRA})}, 2023, pp. 10\,553--10\,559.

\bibitem{jeong_parameterized_2024}
H.~J. Jeong, Z.~Gong, S.~Bansal, and S.~Herbert, ``\BIBforeignlanguage{en}{Parameterized fast and safe tracking ({FaSTrack}) using {DeepReach}},'' in \emph{\BIBforeignlanguage{en}{Learn. {Dyn}. {Control} {Conf}. ({L4DC})}}.\hskip 1em plus 0.5em minus 0.4em\relax PMLR, 2024, pp. 1006--1017.

\bibitem{nakamura_online_2023}
K.~Nakamura and S.~Bansal, ``Online {Update} of {Safety} {Assurances} {Using} {Confidence}-{Based} {Predictions},'' in \emph{{IEEE} {Int}. {Conf}. {Robot}. {Autom}. ({ICRA})}, 2023, pp. 12\,765--12\,771.

\bibitem{ramesh_kumar_fast_2023}
A.~Ramesh~Kumar, K.-C. Hsu, P.~J. Ramadge, and J.~F. Fisac, ``Fast, {Smooth}, and {Safe}: {Implicit} {Control} {Barrier} {Functions} {Through} {Reach}-{Avoid} {Differential} {Dynamic} {Programming},'' \emph{IEEE Control Syst. Lett.}, vol.~7, pp. 2994--2999, 2023.

\bibitem{bajcsy_efficient_2019}
A.~Bajcsy, S.~Bansal, E.~Bronstein, V.~Tolani, and C.~J. Tomlin, ``An {Efficient} {Reachability}-{Based} {Framework} for {Provably} {Safe} {Autonomous} {Navigation} in {Unknown} {Environments},'' in \emph{{IEEE} {Conf}. {Decis}. {Control} ({CDC})}, 2019, pp. 1758--1765.

\bibitem{lafmejani_nmpc-lbf_2022}
A.~S. Lafmejani, S.~Berman, and G.~Fainekos, ``{NMPC}-{LBF}: {Nonlinear} {MPC} with {Learned} {Barrier} {Function} for {Decentralized} {Safe} {Navigation} of {Multiple} {Robots} in {Unknown} {Environments},'' in \emph{{IEEE}/{RSJ} {Int}. {Conf}. {Intell}. {Robots} {Syst}. ({IROS})}, 2022, pp. 10\,297--10\,303.

\bibitem{jacquet_n-mpc_2024}
M.~Jacquet and K.~Alexis, ``N-{MPC} for {Deep} {Neural} {Network}-{Based} {Collision} {Avoidance} exploiting {Depth} {Images},'' in \emph{{IEEE} {Int}. {Conf}. {Robot}. {Autom}. ({ICRA})}, 2024, pp. 13\,536--13\,542.

\bibitem{salzmann_real-time_2023}
T.~Salzmann, E.~Kaufmann, J.~Arrizabalaga, M.~Pavone, D.~Scaramuzza, and M.~Ryll, ``Real-{Time} {Neural} {MPC}: {Deep} {Learning} {Model} {Predictive} {Control} for {Quadrotors} and {Agile} {Robotic} {Platforms},'' \emph{IEEE Robot. Autom. Lett. (RA-L)}, vol.~8, no.~4, pp. 2397--2404, 2023.

\bibitem{barron_bellman_1989}
E.~N. Barron and H.~Ishii, ``The {Bellman} equation for minimizing the maximum cost,'' \emph{Nonlinear Analysis: Theory, Methods \& Applications}, vol.~13, no.~9, pp. 1067--1090, 1989.

\bibitem{kerrigan_invariant_2000}
E.~Kerrigan and J.~Maciejowski, ``Invariant sets for constrained nonlinear discrete-time systems with application to feasibility in model predictive control,'' in \emph{{IEEE} {Conf}. {Decis}. {Control} ({CDC})}, vol.~5, 2000, pp. 4951--4956 vol.5.

\bibitem{liniger_viability_2015}
A.~Liniger and J.~Lygeros, ``A viability approach for fast recursive feasible finite horizon path planning of autonomous {RC} cars,'' in \emph{Int. {Conf}. {Hybrid} {Syst}.: {Comput}. {Control}}.\hskip 1em plus 0.5em minus 0.4em\relax Association for Computing Machinery, 2015, pp. 1--10.

\bibitem{blanchini_survey_1999}
F.~Blanchini, ``Survey paper: {Set} invariance in control,'' \emph{Automatica}, vol.~35, no.~11, pp. 1747--1767, 1999.

\bibitem{ha_hypernetworks_2017}
D.~Ha, A.~M. Dai, and Q.~V. Le, ``\BIBforeignlanguage{en}{{HyperNetworks}},'' in \emph{\BIBforeignlanguage{en}{Int. {Conf}. {Learn}. {Represent}. ({ICLR})}}, 2017.

\bibitem{galanti_modularity_2020}
T.~Galanti and L.~Wolf, ``On the {Modularity} of {Hypernetworks},'' in \emph{Adv. {Neural} {Inf}. {Process}. {Syst}.}, vol.~33, 2020, pp. 10\,409--10\,419.

\bibitem{andersson_casadi_2019}
J.~A.~E. Andersson, J.~Gillis, G.~Horn, J.~B. Rawlings, and M.~Diehl, ``\BIBforeignlanguage{en}{{CasADi}: a software framework for nonlinear optimization and optimal control},'' \emph{\BIBforeignlanguage{en}{Mathematical Programming Computation}}, vol.~11, no.~1, pp. 1--36, 2019.

\bibitem{lin_generating_2023}
A.~Lin and S.~Bansal, ``Generating {Formal} {Safety} {Assurances} for {High}-{Dimensional} {Reachability},'' in \emph{{IEEE} {Int}. {Conf}. {Robot}. {Autom}. ({ICRA})}, 2023, pp. 10\,525--10\,531.

\bibitem{fisac_bridging_2019}
J.~F. Fisac, N.~F. Lugovoy, V.~Rubies-Royo, S.~Ghosh, and C.~J. Tomlin, ``Bridging {Hamilton}-{Jacobi} {Safety} {Analysis} and {Reinforcement} {Learning},'' in \emph{{IEEE} {Int}. {Conf}. {Robot}. {Autom}. ({ICRA})}, 2019, pp. 8550--8556.

\end{thebibliography}

\end{document}